\documentclass[onecolumn]{tamplate/hexafuture}
\usepackage{fontawesome5}   %
\usepackage{environ}

\usepackage{makecell}
\usepackage{array}
\usepackage{tabularx}

\newcommand{\bench}{EmbodiedLab\xspace}

\makeatletter
\DeclareRobustCommand\onedot{\futurelet\@let@token\@onedot}
\def\@onedot{\ifx\@let@token.\else.\null\fi\xspace}
\makeatother

\newcommand{\ind}{\mathds{1}}   %

\renewcommand{\arraystretch}{1.12}
\newcommand{\tablestyle}{\small\setlength{\tabcolsep}{5pt}}   %
\newcommand{\hl}[1]{\cellcolor{hexasurface}#1}                %
\newcolumntype{L}{>{\raggedright\arraybackslash}X}
\newcolumntype{C}{>{\centering\arraybackslash}X}

\theoremstyle{definition}

\newcommand{\ypred}{\hat{y}}
\newcommand{\ytrue}{y^{*}}

\titleformat*{\paragraph}{\sffamily\bfseries}
\titlespacing*{\paragraph}{0pt}{\parskip}{0.6em}

\usepackage{needspace}

\fancypagestyle{firstpage}{%
    \fancyhf{}%
    \renewcommand{\headrulewidth}{0.5pt}%
    \fancyhead[L]{\hexalogo[25pt]}%
    \fancyhead[R]{\includegraphics[height=25pt]{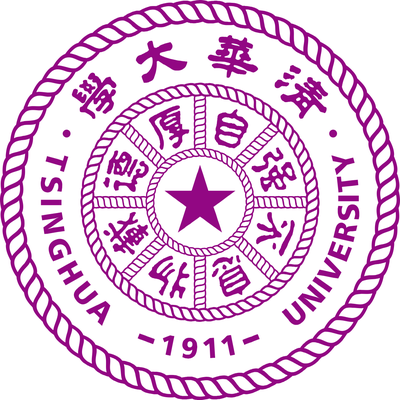}\hspace{8pt}%
      \includegraphics[height=25pt]{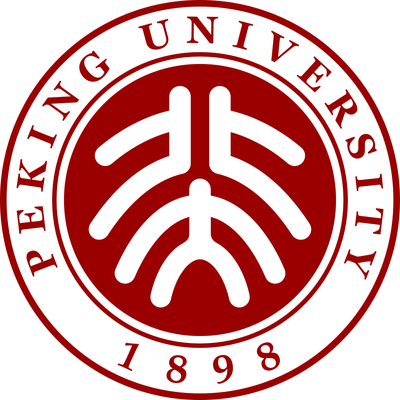}\hspace{8pt}%
      \includegraphics[height=25pt]{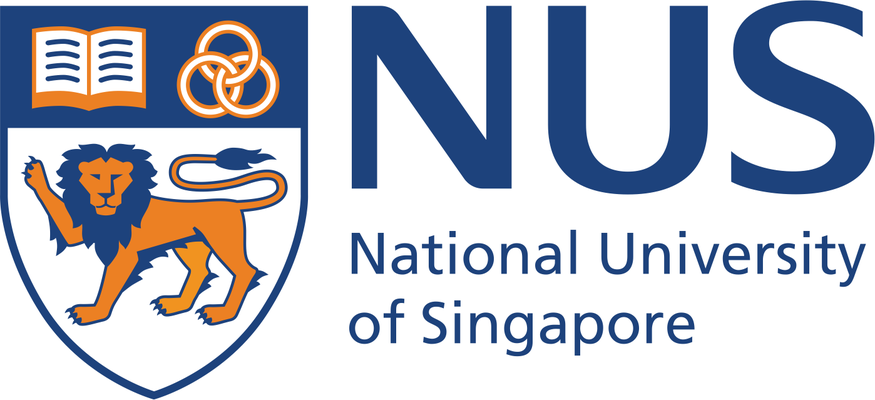}}%
    \fancyfoot[C]{{\sffamily\small\color{hexaprimary}\thepage}}%
}

\RenewEnviron{abstract}{\global\let\abstractbody\BODY}
\def\abstractlist{{\small\color{hexatext}\abstractbody}\vspace{-15pt}}

\newcommand{\maketitlesupplementary}{\appendix\section*{Appendix}\addcontentsline{toc}{section}{Appendix}}



\title{\LARGE EmbodiedLab: Beyond Manipulation, Benchmarking\\ Embodied Agents That Act to Know}

\author{
\begin{center}
    \vspace{-0.7cm}
    {\sffamily\small
    \mbox{Hongcheng Gao\textsuperscript{1,2*\dag}}\hspace{0.9em}
    \mbox{Zelin Zheng\textsuperscript{1*}}\hspace{0.9em}
    \mbox{Alex Zhou\textsuperscript{1}}\hspace{0.9em}
    \mbox{Jay Zhu\textsuperscript{1}}\hspace{0.9em}
    \mbox{Shijia Ge\textsuperscript{1}}\hspace{0.9em}
    \mbox{Yazhe Wang\textsuperscript{1}}\hspace{0.9em}
    \mbox{Xiping Li\textsuperscript{1}}\hspace{0.9em}
    \mbox{Jianshu Zeng\textsuperscript{1,4}}\hspace{0.9em}
    \mbox{Di Wu\textsuperscript{1}}\hspace{0.9em}
    \mbox{Xuan Shangguan\textsuperscript{1}}\hspace{0.9em}
    \mbox{Yexing Wan\textsuperscript{1}}\hspace{0.9em}
    \mbox{Lingyu He\textsuperscript{1}}\hspace{0.9em}
    \mbox{Yongli Wang\textsuperscript{1}}\hspace{0.9em}
    \mbox{Xiaohan Yu\textsuperscript{1}}\hspace{0.9em}
    \mbox{Zhiqi Jia\textsuperscript{1}}\hspace{0.9em}
    \mbox{Yue Liu\textsuperscript{3}}\hspace{0.9em}
    \mbox{Bryan Hooi\textsuperscript{3}}\hspace{0.9em}
    \mbox{Yinpeng Dong\textsuperscript{2}}\hspace{0.9em}
    \mbox{Wentao Zhang\textsuperscript{4}}\hspace{0.9em}
    \mbox{Sihang Wu\textsuperscript{1}}\hspace{0.9em}
    \mbox{Xiao He\textsuperscript{1\dag}}}\\[5pt]
    {\sffamily\footnotesize\color{hexatext}
    \textsuperscript{1}HexaFuture Inc.\quad
    \textsuperscript{2}Tsinghua University\quad
    \textsuperscript{3}National University of Singapore\quad
    \textsuperscript{4}Peking University\\[1pt]
    \textsuperscript{*}Equal contribution\quad
    \textsuperscript{\dag}Corresponding author}\\[4pt]
    {\sffamily\small\bfseries\color{cellblue}\faGlobe\enspace\href{https://hexafuture.ai/benchmarks/embodiedlab/}{Website}}
    \vspace{-0.9cm}
\end{center}
}

\begin{document}
\begin{abstract}
Physical action not only changes the state of an environment but is also the way to know its hidden properties: what cannot be seen must be inferred from the feedback to the change.
As embodied agents advance from executing procedures to pursuing goals, success depends on properties neither given nor visible, so they must act to know.
Existing benchmarks separate changing from knowing: manipulation benchmarks give the target state, so nothing has to be found out, whereas exploration and discovery benchmarks let the agent find out by looking or querying, so nothing physical has to be changed.
Quantitative science experiments offer a natural testbed: the properties sought are new in every instance, appear only in instrument readings under conditions the agent sets up, and can be checked exactly.
We introduce \bench, a simulated laboratory of 35 such experiments.
Given a goal but no procedure, the agent must decide what to measure, operate the apparatus with a robot arm, read instruments through cameras and judge when its readings suffice.
Its 30 determination tasks specify what to find out, whereas its 5 optimization tasks ask only for the best result the apparatus allows.
With native control, the strongest of seven frontier agents succeeds on 41.4\% of the tasks.
Ablations reveal two bottlenecks: (1)~contact-rich manipulation, where ground-truth depth and motion planning change the success rate by at most 10 points, whereas executing every operation for the agent on five of the tasks raises it from at most 20\% to at least 70\%, so what an agent can know is bounded by how reliably it acts; and (2)~closed-loop experimental rigor, where with manipulation supplied, agents fail to gather sufficient readings or read instruments accurately, and most optimization runs stop at one design.
Beyond evaluation, \bench offers an infinitely resamplable environment with verifiable rewards for training embodied agents to explore and understand the physical world.
The benchmark and its evaluation code are available at \url{https://github.com/HexaFuture/EmbodiedLab}.
\end{abstract}

\maketitle
\justifying
\setlength{\parindent}{0pt}

\vspace{-10pt}\noindent\begin{minipage}{\linewidth}
  \centering\includegraphics[width=0.93\linewidth]{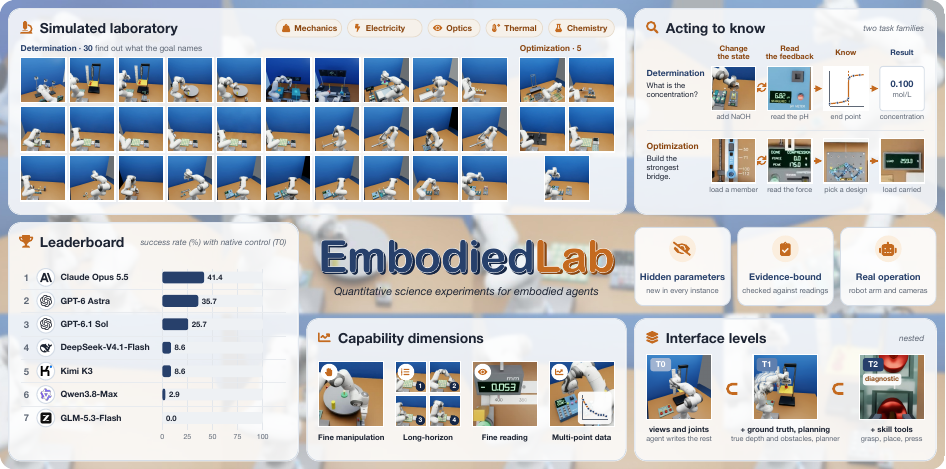}
  \captionof{figure}{\textbf{Overview of \bench.} 35 quantitative experiments in five disciplines: 30 determination tasks ask for what the goal names, and 5 optimization tasks for the best result the apparatus allows. Agents are ranked by success rate with native control (T0).}
  \label{fig:front}
\end{minipage}
\clearpage

\section{Introduction}
\label{sec:intro}

Acting on the physical world not only changes its state: the feedback to that change also reveals properties that observation alone cannot.
A beam's stiffness is invisible at rest; only under a load does the beam bend, and the deflection read on a gauge gives its stiffness.
Executing a given procedure requires only changing the state, whereas pursuing a goal often requires knowing such properties as well.
Embodied agents built on frontier models now write code and call tools to operate robots~\cite{geminirobotics2026,capx2026,selfevolving2026}, solve long-horizon dexterous tasks without task-specific training~\cite{embodiedswe2026} and carry out protocols in real laboratories~\cite{wetrobo2026}.
As they advance from executing procedures to pursuing goals, their success increasingly depends on properties that are neither given nor visible, such as the concentration of a solution or the setting at which a circuit meets its specification.
Such tasks call for \emph{acting to know}: the agent changes the state of the system, observes the feedback, and uses what it learns to complete the task~\cite{kirsh1994,denil2017physics}.

Existing benchmarks separate changing from knowing.
Manipulation benchmarks evaluate \emph{acting to change}: they give the target state and score whether the robot reaches it~\cite{libero2023,robocasa2024,robotwin2025,robodojo2026}, and laboratory robotics benchmarks extend it to instruments and protocol steps~\cite{labutopia2025,autobio2025,pipette2026}; nothing has to be found out.
Embodied question answering asks the agent to find something out, but the answer is already in the scene, and looking suffices~\cite{das2018eqa,gordon2018iqa}.
Scientific discovery benchmarks hide a law or a parameter and let the agent design its experiments~\cite{scienceworld2022,discoveryworld2024,physgym2025,newtonbench2025,activescibench2026}, yet an experiment is a text command or a choice of inputs whose result returns exactly, so nothing physical has to be changed.
In acting to know the two are coupled: how the agent changes the state decides what feedback it can observe, and what it observes decides how it acts next.
Each line of benchmarks removes one side of this coupling, so none can tell whether an agent comes to know by acting (\cref{tab:related}).

\begin{figure*}[t]
  \centering
  \includegraphics[width=\linewidth]{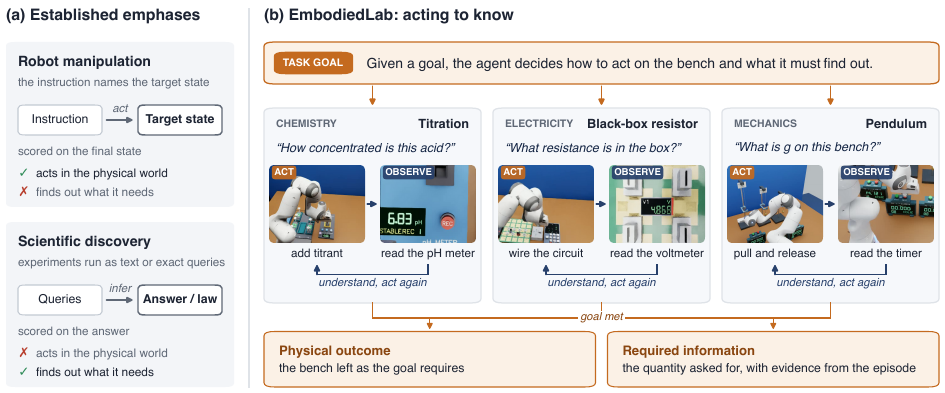}
  \caption{\textbf{Existing benchmark paradigms and acting to know.} (a)~Manipulation benchmarks stop at acting to change the scene into a named target state; science benchmarks find things out without physical action. (b)~Given a goal, an \bench agent changes the state of the apparatus through interaction, observes the feedback and acts again on what it learned; at submission, what it found out, the evidence it recorded and the bench state are scored.}
  \label{fig:teaser}
\end{figure*}

Building such tasks requires three conditions at once: the properties the task depends on are new in every instance, so that nothing can be recalled, they appear only in the feedback to the agent's actions, and what the agent comes to know can be checked exactly.
Quantitative science experiments meet all three: their parameters can be resampled for every instance, show themselves only in instrument readings taken under conditions the agent sets up, and are physical quantities that can be scored exactly.
They are also the established form of acting to know: the experimenter sets the conditions, reads the instruments, repeats the measurement and bounds the error of the result.

We introduce \bench, a simulated laboratory built on Isaac Sim~\cite{isaacsim} in which a robot arm, observed through three cameras, carries out quantitative experiments in mechanics, electricity, optics, thermal physics and chemistry (\cref{fig:front}).
The 30 \emph{determination} tasks specify what to find out and leave to the agent how: which instruments to use, under which conditions to take readings, and when they suffice.
The 5 \emph{optimization} tasks ask for the best result the apparatus allows, such as a bridge that carries the most load or a filter with the strongest attenuation, so the agent must also decide which properties matter, how much each trial is worth and when to stop.
\bench thus evaluates what real-world tasks demand: handed a goal and no procedure, the agent must decide what to measure, operate the apparatus, read the instruments and judge when its readings suffice.

We evaluate seven frontier agents with native control (T0): the strongest, Claude Opus 5.5, succeeds on 41.4\% of the 35 tasks, while four of the seven succeed on fewer than 9\%.
Six milestones (L1--L6) mark where a failed run stops, and interface levels take over parts of the acting; they locate two bottlenecks.
The first is contact-rich manipulation: what an agent can know is bounded by how reliably it acts.
With every operation executed for them, as in science benchmarks, three agents rise from at most 20\% to at least 70\% on five tasks; given the method, none rises.
Yet ground-truth depth and obstacles with motion planning (T1) change the success rate by at most 10 points.
The second is closed-loop experimental rigor.
Given manipulation skills (T2), the weaker agents gain 10 to 27 points, but none reaches the strongest agent; the strongest agents lose 49\% to 56\% of their failed runs after their readings suffice, to a result outside the tolerance or an incomplete experiment; and 80\% of the optimization runs measure at most one design.

\bench is more than a test: it is a playground in which models can acquire the ability it measures.
Frontier models are trained on text, images and video, which contain no experience of acting on a physical system, reading its feedback and being right or wrong about it.
In \bench a model gains this experience first-hand: every reset is a new instance and every episode is scored automatically, so it can collect trajectories of real experimentation without limit and be trained by reinforcement on a reward that guessing cannot earn.
On the optimization tasks it learns to explore the physical world: what to try and when to stop.

Our contributions are summarized as follows.
\begin{itemize}[leftmargin=*,itemsep=2pt,topsep=3pt]
  \item \textbf{A task paradigm: acting to know.} We pose tasks that an agent completes only by interacting with the physical world itself: what the goal depends on is hidden, new in every instance, and appears only in the feedback to action.
  \item \textbf{The \bench benchmark.} 30 determination and 5 optimization tasks from five disciplines in a simulated laboratory, carried out with a robot arm and read through cameras; six milestones and interface levels locate where a run fails.
  \item \textbf{What limits frontier agents, and where to train them.} Contact-rich manipulation limits acting, and agents lack closed-loop experimental rigor even with acting supplied; \bench is also a playground for training both.
\end{itemize}

\section{Related Work}
\label{sec:related}

\noindent\textbf{Embodied manipulation benchmarks.}
RoboCasa~\cite{robocasa2024}, LIBERO~\cite{libero2023}, BEHAVIOR-1K~\cite{behavior1k2024}, RoboTwin~2.0~\cite{robotwin2025} and RoboDojo~\cite{robodojo2026} specify a task by the physical state the scene should reach and score that state.
Later benchmarks lengthen the horizon~\cite{longact2026}, track state transitions~\cite{robograph2026} or add active perception for a manipulation goal~\cite{activearena2026}, and laboratory benchmarks such as LabUtopia~\cite{labutopia2025}, AutoBio~\cite{autobio2025}, SciHorizon-eLab~\cite{scihorizon2026}, Pipette~\cite{pipette2026} and LabDex~\cite{labdex2026} score the completion of protocol steps.
They evaluate acting to change: the target state or procedure is given, and perception serves only to reach it.
In \bench the agent is given a goal but not the state or answer that meets it: it changes the state of the apparatus to know the properties the goal depends on, and it is scored on what it came to know, the evidence it recorded and the state the task requires.

\begin{table*}[t]
  \centering
  \caption{\textbf{Comparison with existing benchmarks.} The goal of a task and what it leaves to the agent, how the agent changes and comes to know the environment, and what is scored.}
  \label{tab:related}
  \tablestyle\footnotesize\setlength{\tabcolsep}{3.1pt}
  \begin{tabular}{lccccc}
    \toprule
    \multirow{2}{*}[-0.4ex]{Benchmark} & \multicolumn{2}{c}{Task} & \multicolumn{2}{c}{Environment} & Metric \\
    \cmidrule(lr){2-3}\cmidrule(lr){4-5}\cmidrule(l){6-6}
    & Goal & Left to the agent & Changing state & Knowing properties & Success \\
    \midrule
    RoboCasa~\cite{robocasa2024} & target state & motion & robot & \textcolor{gray}{not needed} & final state \\
    RoboTwin~2.0~\cite{robotwin2025} & target state & motion & robot & \textcolor{gray}{not needed} & final state \\
    RoboDojo~\cite{robodojo2026} & target state & motion & robot & \textcolor{gray}{not needed} & final state \\
    BEHAVIOR-1K~\cite{behavior1k2024} & goal predicates & action order & robot & scene search & goal predicates \\
    LabUtopia~\cite{labutopia2025} & protocol & motion & robot & \textcolor{gray}{not needed} & steps completed \\
    CaP-X~\cite{capx2026} & target state & control code & robot & \textcolor{gray}{not needed} & final state \\
    \midrule
    EQA~\cite{das2018eqa} & question & where to look & navigation & scene view & answer \\
    ScienceWorld~\cite{scienceworld2022} & science task & experiments & text commands & text feedback & answer, score \\
    DiscoveryWorld~\cite{discoveryworld2024} & science task & experiments & text commands & text feedback & answer, score \\
    PhysGym~\cite{physgym2025} & hidden law & queries & input values & exact outputs & law, value \\
    NewtonBench~\cite{newtonbench2025} & hidden law & queries & input values & exact outputs & law, value \\
    \midrule
    \rowcolor{hexasurface}\textbf{\bench (ours)} & experimental goal & what/how to find out & robot & instrument readings & state, answer, evidence \\
    \bottomrule
  \end{tabular}
\end{table*}

\noindent\textbf{Exploration and scientific discovery benchmarks.}
ScienceWorld~\cite{scienceworld2022} and DiscoveryWorld~\cite{discoveryworld2024} let an agent run experiments through text actions in a symbolic world, and PhysGym~\cite{physgym2025}, NewtonBench~\cite{newtonbench2025} and ActiveSciBench~\cite{activescibench2026} let it set controlled variables and observe the outputs.
Optimization over experiments is the core of self-driving laboratories, where Bayesian optimization and tool-using agents choose the next experiment under cost and noise and automated instruments execute it~\cite{sdlsurvey2026,coscientist2023}, and BuildArena~\cite{buildarena2025} asks language models to design structures that hold under simulated physics.
They evaluate experimental design without embodied execution: the agent issues a text command, an instrument call, a choice of inputs or a design, and the result returns as text or as a computed value.
Embodied question answering~\cite{das2018eqa,gordon2018iqa,esibench2026} evaluates finding out by looking, and interactive perception~\cite{bohg2017interactive} estimates a property of an object, such as its mass or articulation, by acting on it directly; \bench extends this to quantities that need a multi-step experiment.
There the properties appear only in the feedback obtained by changing the state of the apparatus: readings exist only on instruments the agent sets up and reads through its cameras, so how it acts decides what it observes, and an optimization task is scored on the apparatus the agent has built or tuned itself.

\noindent\textbf{Agents that act through code.}
Code as Policies~\cite{liang2023cap}, VoxPoser~\cite{huang2023voxposer}, CaP-X~\cite{capx2026}, EmbodiedSWE~\cite{embodiedswe2026} and RHO~\cite{rho2026} have coding agents compose perception and control primitives.
HexaShell~\cite{hexashell2026} lets an unmodified coding agent drive a dual-arm robot and write its own tools, Physical Coding~\cite{selfevolving2026} refines such tools from verified execution traces, URAI~\cite{urai2026} and PyRUA-Lean~\cite{pyrua2026} let agents write or compose reusable robot tools, coding agents have adapted a robot kit across real wet labs~\cite{wetrobo2026}, and EngiWorld~\cite{engiworld2026} has them operate professional engineering software.
These works are evaluated on manipulation or engineering benchmarks or demonstrated on selected tasks, among them physics experiments on a real robot~\cite{selfevolving2026}.
\bench evaluates them systematically on acting to know: properties are new in every instance, results are scored automatically, and the code must not only move the robot but also take, read and analyze the instrument readings.

\section{\bench}
\label{sec:benchmark}

\subsection{Acting to know as a task}
\label{sec:formulation}

An instance of a task is a physical system whose task-relevant properties $\theta$ are drawn anew and hidden from the agent.
The agent starts with a task description $I_0$ and an initial observation $o_0$; each of its actions changes the state $s_t$ of the system, and the feedback it observes is
\begin{equation}
  o_t=h(s_t,\theta).
  \label{eq:obs}
\end{equation}
An episode $\tau$ succeeds when $\mathrm{succ}(\tau;\theta)=1$.
The task calls for acting to know when three conditions hold: (i)~success depends on $\theta$, so the episodes that succeed differ across instances; (ii)~$\theta$ is neither given in $I_0$ nor visible in $o_0$; and (iii)~the feedback to the agent's actions carries information about $\theta$.
In acting to change, success is $\ind[\,s_\tau\in S_{\mathrm{goal}}\,]$ with the target state $S_{\mathrm{goal}}$ given in $I_0$, and nothing has to be found out.

The two task families differ in what $I_0$ specifies.
In a determination task it names the quantity to find out: the agent submits an estimate $\ypred$ of
\begin{equation}
  \ytrue=q(\theta,\,s_{\tau}),
  \label{eq:target}
\end{equation}
a property of the system, a response it will show, or a configuration that meets a specification; the last two depend on the state $s_\tau$ the agent leaves and are measured in a physical continuation run after the submission.
In an optimization task it names only an objective $J(x;\theta)$ over the configurations $x$ the apparatus allows.
The best achievable value $\max_x J(x;\theta)$ depends on $\theta$, but which parts of $\theta$ matter, how precisely they must be known and how many trials to spend on them are left to the agent, and it is scored on the configuration it actually builds.

\begin{figure*}[t]
  \centering
  \includegraphics[width=\linewidth]{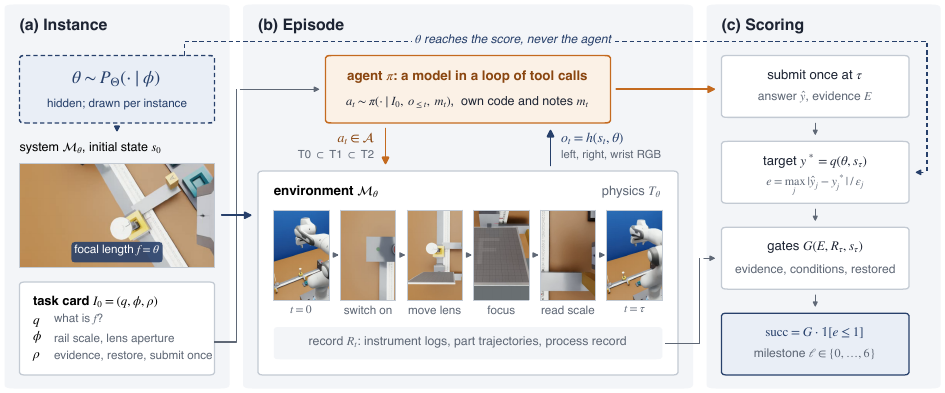}
  \caption{\textbf{Instance, episode and scoring of an \bench task.} (a)~A hidden $\theta$ fixes the system; the task card gives the objective, public parameters and rules, but no method. (b)~The agent acts and observes through three cameras while the environment keeps a record. (c)~Success requires every gate and an error within tolerance of the hidden target.}
  \label{fig:overview}
\end{figure*}

Quantitative science experiments meet the three conditions by construction.
$\theta$ collects the physical parameters of the apparatus and is resampled for every instance; it appears only in instrument readings under conditions the agent sets up; and $\ytrue$ and the best achievable $J$ are computed from $\theta$, so they are hidden from the agent and scored exactly.
An episode follows a single-submission protocol with verifiable evidence.
The agent receives $I_0$ as a task card with the objective, the public parameters and the rules, but no method.
It submits one result together with its evidence $E$, the instrument records the result rests on, which the environment verifies against its own record $R_\tau$ of the instrument logs, the part trajectories and the robot's process (\cref{fig:overview}).

\subsection{Determination and optimization tasks}
\label{sec:tasks}

The 30 determination tasks are textbook experiments in five disciplines, or direct extensions of them (\cref{tab:stats}).
Most ask the agent to measure a property, such as a focal length; others to predict a response, such as where a launched ball will land, or to configure the apparatus to a specification.
Even when what to find out is given, how to find it out is not: the agent chooses the instruments, the conditions and the number of readings and decides when they suffice.
The tasks fall into two tiers: an experiment is \emph{basic} when a standard design fixed in advance meets the tolerance, so that the apparatus dictates the procedure, and \emph{advanced} when the agent must choose the conditions, decide each next condition from the readings so far, identify a model or act on its result (\cref{sec:details}).
The tasks also cover four capability dimensions of difficult interaction: fine manipulation and contact, long-horizon multi-stage work, fine reading, and multi-point acquisition and analysis~\cite{robodojo2026} (\cref{tab:design}).

The 5 optimization tasks ask for the best result the apparatus allows, from the load a truss bridge carries to the efficiency with which a laser is coupled into a fiber (\cref{sec:optim}).
Each delivers a physical artifact and is scored on its actual state at submission.
The apparatus departs from the textbook model in hidden ways, so the textbook design is a starting point that typically falls short; every trial costs simulated time and some destroy material, so what to try, how often and when to stop are themselves decisions.
Four also require the agent to characterize what it has built, by a reading or a prediction of its own result that is scored for accuracy.

\subsection{An open laboratory}
\label{sec:lab}

\begin{figure*}[p]
  \centering
  \includegraphics[width=\linewidth]{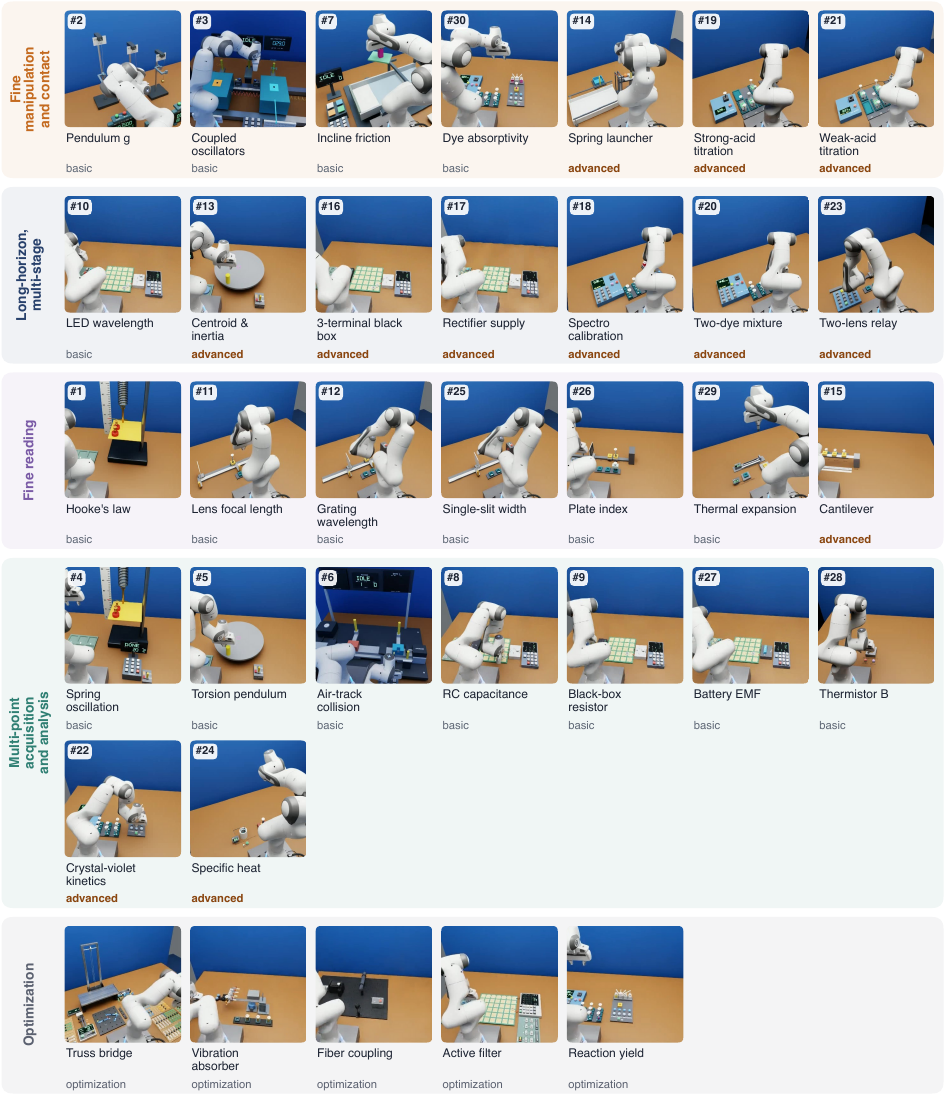}
  \caption{\textbf{Overview of the 35 tasks.} The 30 determination tasks, grouped by primary capability dimension, and the 5 optimization tasks are shown as seen by one of the robot's fixed cameras during a reference run. The tag gives the tier or the family.}
  \label{fig:gallery}
\end{figure*}

Every bench is a parameterized scene in Isaac Sim~\cite{isaacsim} with a Franka Panda arm, a two-finger gripper and three cameras, two fixed and one on the wrist; circuits, optics, heat flow, spectra and solution chemistry are computed by standard physical models and rendered back into the scene (\cref{fig:gallery}, \cref{sec:details}).
Benches carry more instruments, tools and materials than one method needs, instruments show quantized and noisy values on scales, pointers and displays that the agent reads through its cameras, and every operation is physical.

\begin{figure*}[tp]
  \centering
  \includegraphics[width=\linewidth]{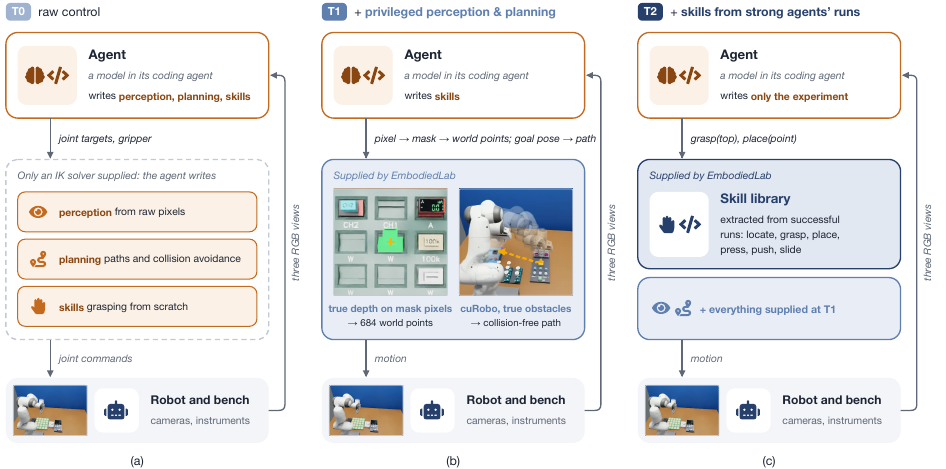}
  \caption{\textbf{Interface levels.} Robot, bench and camera views are the same at every level; each level supplies more between the agent and the robot. (a)~T0 supplies only the camera images, the robot state, joint position targets, gripper commands and an inverse-kinematics solver that checks no collisions and plans no path. (b)~T1 supplies ground-truth depth on segmentation masks and cuRobo~\cite{curobo} planning against the true obstacles, shown with real outputs of T1 runs. (c)~T2 adds a skill library, extracted from the successful runs of the strongest agents, in the agent's workspace.}
  \label{fig:levels}
\end{figure*}

The agent acts through a programming interface in its workspace, on which it can build its own tools, at three nested levels (\cref{fig:levels}, \cref{tab:exposure}).
T0 is native control: the interface returns the camera images with their calibration and the robot state and moves the robot only by joint position targets and gripper commands, so trajectories, collision avoidance, perception and manipulation are the agent's own.
T1 adds a promptable segmentation model~\cite{sam3} with ground-truth depth on its masks and collision-free Cartesian motion planning~\cite{curobo} on the ground-truth obstacles.
T2 adds skills for locating, grasping, carrying, placing, pressing, pushing and sliding, extracted from the successful runs of the strongest agents as short sequences of public T1 calls.
Every T0 policy is also a T1 and a T2 policy, so each level acts as an intervention that takes over part of the acting while the agent still reads, records and analyzes for itself.
T0 and T1 form the benchmark. T2 is a diagnostic condition for the four agents that lose most of their runs before their readings suffice: it asks how much of their failure remains once manipulation is supplied, and how far they then carry the experiment.

\subsection{Evaluation}
\label{sec:evaluation}

At the submission the environment seals the record $R_\tau$, and the run is scored as
\begin{equation}
  e=\max_{j}\frac{d_j}{\epsilon_j},
  \qquad
  \mathrm{succ}=G(E,R_\tau,s_\tau)\cdot\ind[\,e\le 1\,],
  \label{eq:score}
\end{equation}
where $d_j$ is the error of scored quantity $j$ and $\epsilon_j$ its threshold, and $G\in\{0,1\}$ is the conjunction of gates evaluated on the sealed record and the final state.
In a determination task $d_j=|\ypred_j-\ytrue_j|$ and $\epsilon_j$ is the tolerance set by the instruments; in an optimization task the main term is the shortfall $\max(0,\,J_{\mathrm{ref}}-J(x;\theta))$ of the built configuration $x$ from a reference value $J_{\mathrm{ref}}$ computed from $\theta$, and a reported reading adds a term for its error.
The gates check that every cited reading was produced by an instrument in this episode under the required conditions, that the reported configuration matches the actual one, and that the bench is restored, with every part the objective does not use back in its place.
In acting to know, the changes the agent makes are means, not the goal: a finished task leaves only the changes the objective asks for, so the agent must keep track of what it changed and undo the rest.

To locate where a failed run stops (\cref{sec:where}), each run is checked against six nested milestones: the robot acts on the apparatus (L1); the apparatus is in a state from which a reasonable method can measure (L2); a first valid reading is taken (L3); the valid readings suffice for every scored quantity (L4); every gate except restoring the bench holds and $e\le 1$ (L5); and the bench is restored, so that L6 is success.
L1--L4 are minimal checks for each apparatus that any valid method passes (\cref{sec:cards}); they do not prescribe a method.
In an optimization task L2--L4 concern designs instead of readings: a design is ready to be evaluated (L2), its first result is obtained (L3), and two different designs have results (L4).
L5 is a valid design whose reported value, where one is asked for, is accurate, and L6 a design within the threshold of the reference; restoring the bench is credited separately (\cref{sec:setup}).

\subsection{Construction and validation}
\label{sec:construction}

Every task passes five steps before release: an apparatus, a declarative specification, instances generated from seeds, a reference solution that fixes the tolerance or success threshold, and a release audit (\cref{fig:build}).
Each task is solvable: a reference solution that acts through the public interface and takes every reading from the instruments, without the hidden parameters, meets the tolerances of all 30 determination tasks, the tightest the instruments support (\cref{sec:details}), and reaches the success threshold of the objective on every optimization instance it is calibrated on (\cref{tab:stats}).
Each task also needs experimenting: across instances the target of a determination task ranges over at least 6 times its tolerance and 90 times at the median, so that the best single answer given without an experiment meets the tolerance in 4\% of instances on average (\cref{sec:details}), and designs derived from the public parameters and the textbook model reach the objective threshold on 6\% of the instances (\cref{sec:optim}).

\section{Experiments}
\label{sec:experiments}
\suppressfloats[t]  %

\begin{table*}[t]
  \centering
  \caption{\textbf{Performance of seven frontier agents on \bench with native control (T0).} Each entry is the success rate (\%) with the process score (0--100) in gray. The first column, by which the agents are ranked, covers all 35 tasks with every task weighted equally; the other columns give each task family and tier. Attainment is the mean achieved objective relative to the best known value. Tokens per run are averaged over all 35 tasks (\cref{tab:cost}). Numbers of tasks are in parentheses and the best success rate is in bold.}
  \label{tab:main}
  \tablestyle\setlength{\tabcolsep}{4pt}
  \newcommand{\ps}[1]{\,{\scriptsize\color{gray}#1}}
  \begin{tabular}{@{}lcccccccc@{}}
    \toprule
    & & \multicolumn{3}{c}{Determination} & \multicolumn{2}{c}{Optimization (5)} & \multicolumn{2}{c}{Tokens per run} \\
    \cmidrule(lr){3-5}\cmidrule(lr){6-7}\cmidrule(l){8-9}
    Agent & All (35) & All (30) & Basic (18) & Advanced (12) & Success rate & Attainment & Input (M) & Output (k) \\
    \midrule
    Claude Opus 5.5~\cite{opus552026} & \textbf{41.4}\ps{64.4} & \textbf{43.3}\ps{64.0} & 38.9\ps{61.9} & \textbf{50.0}\ps{67.1} & \textbf{30.0}\ps{67.0} & \textbf{0.55} & 31.5 & 151.4 \\
    GPT-6 Astra~\cite{gpt6astra2026} & 35.7\ps{62.5} & 40.0\ps{62.8} & \textbf{50.0}\ps{73.9} & 25.0\ps{46.2} & 10.0\ps{61.0} & 0.30 & 20.2 & 31.3 \\
    GPT-6.1 Sol~\cite{gpt61sol2026} & 25.7\ps{53.9} & 28.3\ps{55.7} & 36.1\ps{58.9} & 16.7\ps{50.8} & 10.0\ps{43.0} & 0.26 & 33.5 & 53.2 \\
    DeepSeek-V4.1-Flash~\cite{deepseekv41flash2026} & 8.6\ps{35.1} & 10.0\ps{36.0} & 16.7\ps{38.9} & 0.0\ps{31.7} & 0.0\ps{30.0} & 0.08 & 52.3 & 257.4 \\
    Kimi K3~\cite{kimik32026} & 8.6\ps{32.9} & 10.0\ps{35.7} & 16.7\ps{38.3} & 0.0\ps{31.7} & 0.0\ps{16.0} & 0.03 & 102.5 & 301.8 \\
    Qwen3.8-Max~\cite{qwen38max2026} & 2.9\ps{17.5} & 3.3\ps{16.3} & 5.6\ps{20.6} & 0.0\ps{10.0} & 0.0\ps{25.0} & 0.07 & 33.5 & 523.2 \\
    GLM-5.3-Flash~\cite{glm53flash2026} & 0.0\ps{16.3} & 0.0\ps{16.3} & 0.0\ps{18.9} & 0.0\ps{12.5} & 0.0\ps{16.0} & 0.00 & 60.6 & 243.7 \\
    \bottomrule
  \end{tabular}
\end{table*}

\subsection{Setup}
\label{sec:setup}

\noindent\textbf{Models.}
We evaluate seven frontier models: GPT-6 Astra~\cite{gpt6astra2026}, GPT-6.1 Sol~\cite{gpt61sol2026}, Claude Opus 5.5~\cite{opus552026}, DeepSeek-V4.1-Flash~\cite{deepseekv41flash2026}, Kimi K3~\cite{kimik32026}, GLM-5.3-Flash~\cite{glm53flash2026} and Qwen3.8-Max~\cite{qwen38max2026}. Each acts as the agent in a loop of tool calls: it observes the bench through the three cameras and operates the robot through the interface of the level. Every model runs at the default settings of its vendor, including the reasoning effort (\cref{sec:prereg}).

\noindent\textbf{Protocol.}
Every agent is evaluated on the determination tasks at T0 and T1 and on the optimization tasks at T0 and T1, with the same task card, budget and wall-clock limit at every level, and submits once per run; the DeepSeek, Kimi, GLM and Qwen agents are also run under the diagnostic condition T2 (\cref{sec:lab}).
A run that does not succeed counts as a failure, whether the agent stops without submitting, exhausts the simulated-time or wall-clock budget, or submits a result that lies outside the tolerance or fails a gate (\cref{sec:prereg}).

\noindent\textbf{Metrics.}
A run either succeeds (\cref{eq:score}) or ends at the highest milestone it passes together with all before it (\cref{sec:evaluation}).
Its process score, from 0 to 100, credits the progress of a failed run: each milestone it passes adds a weight (\cref{sec:prereg}).
An optimization run is also measured by its \emph{attainment} $J/\bar{J}$, the achieved objective relative to the best value known for the instance (\cref{sec:optim}).
Success rates average first over the runs of each task and then over the tasks, so that every task counts equally; process scores and attainments are averaged in the same way.
The headline metric is the success rate over all 35 tasks at T0.

\subsection{Can frontier agents act to know?}
\label{sec:results}

Frontier agents cannot yet act to know reliably.
The strongest, Claude Opus 5.5, succeeds on 41.4\% of the 35 tasks at T0, followed by GPT-6 Astra at 35.7\% and GPT-6.1 Sol at 25.7\%; the DeepSeek, Kimi, Qwen and GLM agents succeed on at most 8.6\% (\cref{tab:main}).
The process scores order the agents in the same way, from 64.4 to 16.3, so the failed runs of the stronger agents also get further.

Leaving more to the agent separates them further: six of the seven agents succeed less often on advanced than on basic experiments, four of them on none; only Claude Opus 5.5 succeeds more often (50.0\% against 38.9\%).
The optimization tasks leave the most to the agent, which must also decide what to try, how often and when to stop, and no agent succeeds more often on them than on the determination tasks: the success rate is 30.0\% for Claude Opus 5.5, 10.0\% for GPT-6 Astra and for GPT-6.1 Sol, and 0\% for the other four, and the mean attainment is 0.55, 0.30 and 0.26 for the first three and at most 0.08 for the rest.

\subsection{What limits them?}
\label{sec:insights}

\begin{figure*}[t]
  \centering
  \includegraphics[width=\linewidth]{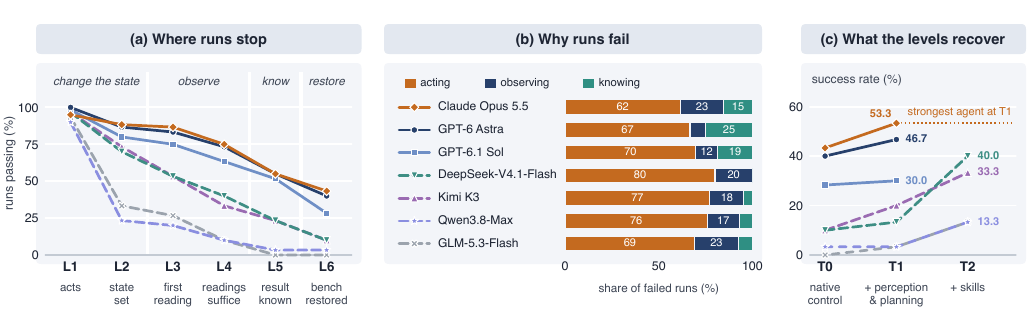}
  \caption{\textbf{Diagnosis by milestone, failure stage and interface level.} (a)~Runs at T0 that pass each milestone; the last point is the success rate. (b)~Failed runs at T0 by the stage of their first decisive error; the rows give the key to the agents. (c)~Success rate at T0, at T1 and, for four agents, under the diagnostic condition T2.}
  \label{fig:ladder}
\end{figure*}

We take the coupling apart in three ways (\cref{fig:ladder}): by milestone, by the first decisive error of each failed run, the earliest one the agent does not correct and after which success is impossible (\cref{sec:causes}), and by level.

\noindent\textbf{Strong agents lose their runs after their readings suffice, weak agents before.}\label{sec:where}
Among the runs that reach the previous milestone, the three strongest agents pass each of L1 to L4 at rates from 0.81 to 1.00.
Their lowest rates are at knowing the result within tolerance (L5, 0.73 to 0.82) and at restoring the bench (L6, 0.55 to 0.79; \cref{fig:ladder}a).
These last steps leave no room for error: the agent submits once, and a single misread digit puts the result outside the tolerance.
As a result, 49\% to 56\% of their failed runs end after the readings suffice.
The other four agents lose 60\% to 90\% of their runs earlier, while changing the state of the apparatus or observing the feedback.
On the determination tasks the budget ends only 8.6\% of the failed runs; the others submit an answer or stop with budget left (\cref{sec:sensitivity}).

\noindent\textbf{Acting errors dominate the failures of every agent.}\label{sec:code}
With causes grouped by the stages of acting to know, acting on the apparatus is the leading cause of failure for every agent, at 62\% to 80\% of its failed runs, and knowing accounts for at most 25\% (\cref{fig:ladder}b; causes and examples in \cref{fig:failures}).
For the DeepSeek, Kimi, GLM and Qwen agents contact alone accounts for 45\% to 62\%: grasps that slip, parts that tip over, keys that are missed.
For the three strongest agents contact falls to 19\% to 28\%; the rest of their acting failures are collisions of the arm and parts set down out of place.
Observing errors are rarer but final, as reading instruments is still hard for vision-language models~\cite{measurebench2025}: one misread digit decides the run, and they account for 8\% to 23\% of each agent's failed runs, the second largest share for five of the seven.
Precision is rarely missing: on the determination tasks a tenfold tolerance raises the mean success rate only from 19.3\% to 20.7\% and leaves the ranking of the agents unchanged (\cref{sec:sensitivity}).

\noindent\textbf{In acting, the limit is contact-rich manipulation, not 3D perception or motion planning.}\label{sec:boundary}
T1 adds ground-truth depth and obstacles with collision-free motion planning, and the success rate changes by 0 to 10.0 points (\cref{fig:ladder}c): planned motion avoids collisions but leaves the grasp and the placement to the agent, where the runs fail.
The same holds on the optimization tasks: the success rate changes by 0 to 10.0 points, and the four agents at 0\% stay there (\cref{tab:crossopt}).
The mean attainment of every agent changes by at most 0.08, the largest change being from 0.55 to 0.63.

\noindent\textbf{Supplied skills lift the weaker agents, but not to the level of the strongest.}
The diagnostic condition T2 supplies manipulation skills to the four agents that lose most runs before their readings suffice.
Their success rates rise by 10.0 to 26.7 points, and the best of the four reaches the T0 success rate of GPT-6 Astra (\cref{fig:ladder}c, \cref{tab:cross}).
On the optimization tasks the DeepSeek and Kimi agents rise from 0\% to 10.0\%, and the Qwen and GLM agents stay at 0\% (\cref{tab:crossopt}).
Both successes are on the reaction task, and the mean attainment of the four agents stays at or below 0.18, against 0.55 for Claude Opus 5.5 at T0.
In the experiment itself a gap remains: the four agents complete their readings (L4) in 40\% to 77\% of runs, against 90\% for Claude Opus 5.5 at T1, and none reaches its success rate.

\noindent\textbf{Most optimization runs test at most one design.}\label{sec:explore}
An optimization task can be solved only by building alternative designs and comparing their measured results.
We compare each run with the textbook design, which needs no experiment, and with the best result the apparatus allows (\cref{sec:optim}).
In 66\% of the runs the agent never measures a design of its own: blocks are not seated, lenses not mounted, weights not placed.
On the four tasks where a run can measure several designs, only 25\% of the runs do, and no run on fiber coupling compares two lenses; in all, 80\% of the runs measure at most one design.
Agents also stop early: of the unsuccessful runs that the agent ends itself, 51\% leave more than a third of the budget.
Only 20\% of the runs end above the textbook design, and every successful run departs from it.

\begin{table}[t]
  \centering
  \caption{\textbf{Ablation of acting, method and reading.} Success rate (\%) of three agents. A tick marks a part left to the agent, and the first row is the unchanged task at T0. Acting is taken over by tool calls that execute every operation and always succeed, the method by a handout with the procedure and its formulas, and reading by the readings given as text. The conditions on the bench are run on five representative tasks, one from each discipline; the last needs no bench and is run on most determination tasks. ``As in'' names the kind of benchmark that leaves the same parts to the agent.}
  \label{tab:ablation}
  \tablestyle\footnotesize\setlength{\tabcolsep}{3.4pt}
  \newcommand{\yes}{\checkmark}
  \newcommand{\no}{\textcolor{gray}{--}}
  \begin{tabularx}{0.62\linewidth}{@{}cccLccc@{}}
    \toprule
    \multicolumn{3}{@{}c}{Left to the agent} & & \multicolumn{3}{c@{}}{Success rate (\%)} \\
    \cmidrule{1-3}\cmidrule{5-7}
    Acting & Method & Reading & As in & \makecell{GPT-6\\Astra} & \makecell{DeepSeek-\\V4.1-Flash} & \makecell{Kimi\\K3} \\
    \midrule
    \hl{\yes} & \hl{\yes} & \hl{\yes} & \hl{\bench} & \hl{20} & \hl{0} & \hl{0} \\
    \yes & \no & \yes & manipulation & 20 & 0 & 0 \\
    \no & \yes & \yes & science & 90 & 80 & 70 \\
    \no & \no & \yes & visual QA & 80 & 70 & 40 \\
    \no & \no & \no & computation & 96 & 96 & 100 \\
    \bottomrule
  \end{tabularx}
\end{table}

\noindent\textbf{Agents can find out once the acting is done for them.}\label{sec:ablation}
Existing benchmarks evaluate one side of acting to know (\cref{tab:related}).
We remove one side on five of our own tasks, one from each discipline (Hooke's law, specific heat, lens focal length, black-box resistor and strong-acid titration), and change nothing else (\cref{tab:ablation}, \cref{sec:ablationsetup}).
With the method and formulas given, the agent only has to act, as in a manipulation benchmark: the success rates of GPT-6 Astra and the DeepSeek and Kimi agents stay at 20\%, 0\% and 0\%.
With every operation executed for it, the agent only has to find out, as in a science benchmark: the success rates rise to 90\%, 80\% and 70\%.
With both, what is left is reading the instruments, yet the rates fall to 80\%, 70\% and 40\%, below those of agents that plan the experiment themselves.
They act and observe as often as without the handout, but follow its steps without adapting to the transient state of the bench, such as reading a pointer before the tray comes to rest: a given procedure does not supply closed-loop rigor.
With the readings given as text, which needs no bench and is run on 27 determination tasks, the agents compute the answer in 96\% to 100\% of them, against 0\% to 48\% on the same tasks at T0.
The agents have the physics and can use the readings.
What they cannot do is produce the readings by their own actions.
Once the operations are executed, the remaining failures are in observing and knowing and fall on two tasks: on the lens a position is read from an image out of focus, and on the specific heat the heat loss is left uncorrected (\cref{tab:ablationtasks}).
At T0 the run ends before reaching them.

\noindent\textbf{Both missing abilities can be trained in \bench.}
Contact-rich manipulation limits acting, so agents need reliable grasping and placing rather than more 3D perception or planning.
In \bench they can be trained for it by reinforcement: every episode is scored automatically, milestone by milestone, and the reward comes only when the apparatus ends up as the experiment needs.
Closed-loop experimental rigor is the second limit: agents need to read instruments without a wrong digit, keep experimenting until the evidence suffices, and keep track of every change to the bench.
Every episode leaves a trajectory with its readings and the verified outcome, and every reset is a new instance, so this can be learned from unlimited data; the optimization tasks train what to try and when to stop.
Both limits show only because acting and knowing are evaluated together.

\section{Conclusion}
\label{sec:conclusion}
We introduced \bench, a benchmark of acting to know.
Its 35 quantitative experiments in a physically simulated laboratory give the agent a goal but no procedure: it must decide what to measure, operate the apparatus under contact, read the instruments through cameras and infer the result.
The strongest of seven frontier agents succeeds on 41.4\% of the tasks, and two bottlenecks account for the failures.
The first is contact-rich manipulation: agents succeed once the operations are executed for them but not when given the method, ground-truth depth or motion planning, so what an agent can know is bounded by how reliably it acts.
The second is closed-loop experimental rigor: even with manipulation supplied, agents misread instruments and stop before their readings suffice, and most optimization runs measure one design.
Both can be trained in \bench, where every reset is a new instance and every outcome is verified: trajectories of real experimentation without limit and a reward that guessing cannot earn.
We release \bench as a benchmark and a playground for embodied agents that explore and understand the physical world.

\bibliographystyle{unsrtnat}
\bibliography{main}

\clearpage
\clearpage
\maketitlesupplementary
\makeatletter\setlength{\@fptop}{0pt}\setlength{\@dblfptop}{0pt}\makeatother
\renewcommand{\topfraction}{0.95}\renewcommand{\bottomfraction}{0.9}\renewcommand{\textfraction}{0.05}\renewcommand{\floatpagefraction}{0.8}\renewcommand{\dbltopfraction}{0.95}\renewcommand{\dblfloatpagefraction}{0.8}

\section{Optimization tasks}
\label{sec:optim}

Each optimization task delivers a physical artifact and is scored on its actual state at the submission; the configuration the agent reports is only checked against it (\cref{tab:optimdetail}).
Of the submitted runs, 84\% report the reading of the instrument, yet 43\% misstate the configuration they leave.
The reference $J_{\mathrm{ref}}$ of each instance is computed offline from the hidden parameters and fixed before any evaluation.
On the fiber and absorber tasks it is the exact optimum over the configurations the bench and the robot can realize.
On the bridge it is the strongest design at the nominal properties of the four member materials, solved to within 0.2\%, and on the reaction task the highest yield that the procedure of the reference solution can reach; a run can exceed these two.
The breadboard of the active filter admits far more circuits than can be searched exhaustively, so its success threshold uses the best unity-gain two-stage Sallen--Key filter the components allow, found exactly, while its attainment uses the best circuit found by a search over board layouts, 18 to 26\,dB stronger; a circuit beyond the standard topology thus succeeds and still ranks above it.
The attainment $J/\bar{J}$ ($\bar{J}/J$ for an objective to minimize), with $\bar{J}$ the best value known for the instance, is not capped at one and is zero without a valid design.
The reference solution reaches the success threshold of the objective on every calibrated instance, with an attainment of 0.99 or more on the fiber, reaction and absorber tasks and 0.94 on the bridge; on the filter the best standard circuit reaches 0.59 of the best circuit found on the board.
The bridge is loaded once until it fails, so its milestones L2--L4 are a bridge ready to load, the start of loading and a first valid load reading.

In every task the textbook design is a starting point that typically falls short, because the apparatus departs from the ideal in hidden ways: printed members crush or carry an initial bow, an unseen side reaction consumes the product, an end stop and a nonlinear damper break the linear tuning, the op-amp bandwidth raises the quality factor, and lens aberrations and knob backlash shift the coupling optimum.
Exploration has a cost: member tests are destructive, cuvettes and components are limited, the heating block cools slowly, and every sweep, heating step and grasp is charged to the simulated-time budget.
Instances that need no exploration are excluded: on the reaction and absorber tasks those whose optimum lies at the edge of the adjustable range, where pushing a setting to its limit would be optimal, and on the truss bridge those whose two member materials are too alike for the choice of material to matter.

The textbook design of \cref{sec:explore} is the design that needs no experiment: it is derived from the public information and the standard textbook model alone (\cref{tab:noexp}).
Where the card gives no value for a property, such as the member strengths, the rate constants or the beam and damper constants, it takes the center of the stated range.
It is scored with the true parameters of the instance by the bench's own model, with building, reporting and restoring the bench assumed perfect, so the results bound from above what can be reached without experimenting.
Such designs succeed on 6\% of the instances, with a mean attainment of 0.55 over the tasks.
On the filter the nominal design attenuates as strongly as the best standard circuit, but with the true component values its passband ripple exceeds 0.5\,dB on every instance; designing with the op-amp model of the card instead, or taking a fourth-order Butterworth filter with the nearest nominal values, does no better (attainment 0.53 and 0.32).
On the fiber the mode-matched lens would succeed on 60\% of the instances if the beam were aligned and focused perfectly, but alignment needs measurement, and without it the beam misses the core.
On the reaction, the best fixed conditions we found reach an attainment of 0.81 and succeed on 10\% of the instances, no more often than the textbook design.

\begin{table}[H]
  \centering
  \caption{\textbf{Optimization designs chosen without an experiment.} For each task, the textbook design derived from the public information alone is scored with the true parameters of every instance, assuming a perfect build. Succ.: share of instances on which it succeeds; Att.: mean attainment.}
  \label{tab:noexp}
  \footnotesize\setlength{\tabcolsep}{3pt}
  \begin{tabularx}{0.62\linewidth}{@{}l>{\raggedright\arraybackslash}Xrr@{}}
    \toprule
    Task & Design & Succ. & Att. \\
    \midrule
    Truss bridge & strongest bridge for mid-range member strengths & 20\% & 0.77 \\
    Reaction yield & best conditions for mid-range rate constants & 10\% & 0.78 \\
    Vibration absorber & Den Hartog tuning with nominal beam, blade and damper & 0\% & 0.62 \\
    Active filter & best two-stage Sallen--Key for nominal components and an ideal op-amp & 0\% & 0.59 \\
    Fiber coupling & mode-matched lens, mirrors left at the start & 0\% & 0.00 \\
    \bottomrule
  \end{tabularx}
\end{table}

\begin{table*}[tp]
  \centering
  \caption{\textbf{The 5 optimization tasks.} For each task, the artifact the agent delivers, the hidden properties its quality depends on, the reference and the success threshold. The reference is computed offline from the hidden parameters. Success also requires the reported value, where asked for, within the given error, and attainment is the achieved objective relative to the best known value.}
  \label{tab:optimdetail}
  \tablestyle\footnotesize
  \begin{tabularx}{\linewidth}{@{}l>{\hsize=.8\hsize}L>{\hsize=1.25\hsize}L>{\hsize=.95\hsize}Ll@{}}
    \toprule
    Task & Delivered artifact & Hidden properties & Reference & Success \\
    \midrule
    Truss bridge & a pinned truss of at most 50\,g, loaded with weights until it fails & strength, buckling and crushing of four member materials & strongest buildable bridge $P^{*}$ & peak load $\ge0.9\,P^{*}$ \\
    Reaction yield & one quenched tube of product & rate constants and activation energies of the target, over-reaction, side and dimer reactions & highest yield $y^{*}$ under robot-feasible timing & $y\ge0.98\,y^{*}$; prediction within 5\% \\
    Vibration absorber & absorber tuned by mass, spring length and damper setting & beam modulus, clamp stiffness, spring tolerance, damper curve, end-stop gap & smallest swept peak $A^{*}$ over all settings & peak $\le1.05\,A^{*}$; report within 5\% \\
    Active filter & two-stage low-pass circuit on a breadboard & true component values, op-amp bandwidth and output resistance & worst-case attenuation over 2--20\,kHz of the best two-stage Sallen--Key filter $A_{\mathrm{ref}}$, ripple $\le0.5$\,dB & within 1\,dB of $A_{\mathrm{ref}}$; report within 0.3\,dB \\
    Fiber coupling & beam path through two mirrors and a chosen lens into a single-mode fiber & beam waist, mode-field diameter, lens aberrations, knob backlash and crosstalk & highest coupling efficiency $\eta^{*}$ into the core & $\eta\ge0.95\,\eta^{*}$; report within 2\% \\
    \bottomrule
  \end{tabularx}
\end{table*}

\section{Benchmark details}
\label{sec:details}
\begin{figure*}[tp]
  \centering
  \includegraphics[width=\linewidth]{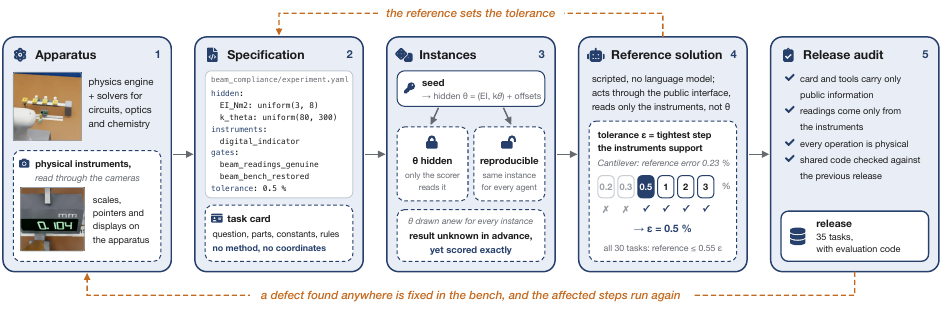}
  \caption{\textbf{Construction pipeline of the \bench benchmark.} Every task passes five steps before release, shown on the cantilever experiment; a defect found at any step is fixed in the bench and the affected steps run again.}
  \label{fig:build}
\end{figure*}

\Cref{tab:stats} summarizes the benchmark.

\noindent\textbf{Physics models.}
Rigid-body mechanics runs in the physics engine at 240\,Hz. What the engine does not model is computed by standard physical models and rendered back into the scene: circuits by modified nodal analysis with exact exponential time stepping, with each LED a Shockley junction whose turn-on energy follows its photon energy; optical images by paraxial ray-transfer matrices with a blur circle and an illuminance that falls with the magnification, grating orders as spots at their diffraction angles and a single slit by its Fraunhofer pattern; heat by lumped bodies that exchange heat with each other and lose it to the room, advanced exactly over each step and read through thermometers with their own lag, with a thermistor that follows the $B$ law and is heated by the meter's test current and a rod whose expansion moves a dial indicator; dye spectra as Gaussian bands fitted to literature spectra and read by a single-beam spectrophotometer with stray light, baseline drift and absorbance-dependent noise, with crystal violet fading at its rate constant; and titrations by an ideal-solution pH solver.

\noindent\textbf{Budgets.}
Every physics step is charged to a budget of simulated time, between 200 and 1500\,s per determination task and between 1900 and 9000\,s per optimization task, and every run has a wall-clock limit.

\begin{table}[t]
  \centering
  \caption{\textbf{Benchmark statistics.} Counts and ranges for the two task families. The counts by discipline follow the order mechanics, electricity, optics, thermal physics, chemistry. Per-task details are in \cref{tab:experiments,tab:optimdetail}.}
  \label{tab:stats}
  \footnotesize\setlength{\tabcolsep}{3pt}
  \begin{tabularx}{0.62\linewidth}{@{}>{\raggedright\arraybackslash}p{2.35cm}LL@{}}
    \toprule
    & Determination & Optimization \\
    \midrule
    Tasks & 30 & 5 \\
    By discipline & 10 / 6 / 5 / 3 / 6 & 2 / 1 / 1 / 0 / 1 \\
    Deliverable & 27 measured, 1 predicted, 2 configured & artifact that reaches the reference \\
    Tiers & 18 basic, 12 advanced & -- \\
    Apparatus platforms & 14 & 5 \\
    Success & error within 0.2\% to 5\% & shortfall within 2\% to 10\% or 1\,dB \\
    Sim.\ time budget & 200 to 1500\,s & 1900 to 9000\,s \\
    Reference solution & solves 30 of 30 & reaches every objective threshold \\
    \bottomrule
  \end{tabularx}
\end{table}

\noindent\textbf{Tiers.}
Let a \emph{measurement design} $\xi=(\xi_1,\dots,\xi_K)$ be the sequence of conditions under which readings are taken, such as the loads on a spring, the positions of a screen or the wavelengths of a spectrometer.
An experiment is \emph{basic} if a standard design $\xi^{\mathrm{std}}$, fixed in advance and independent of $\theta$ and of the readings, together with a standard estimator meets the tolerance: the apparatus dictates the procedure, and the agent has to execute it.
An experiment is \emph{advanced} if no fixed design does, because at least one of four scientific decisions is required:
(a)~\emph{choosing the conditions}, when the precision or the identifiability of $\ytrue$ depends on a $\xi$ the agent must choose;
(b)~\emph{sequential design}, when $\xi_k$ must depend on the readings $y_{<k}$;
(c)~\emph{model identification}, when several parameters are identifiable only jointly or the model structure itself is unknown;
(d)~\emph{acting on the result}, when the deliverable is a prediction or a configuration whose physical continuation is scored, as in \cref{eq:target}.

\noindent\textbf{Same-apparatus pairs.}
Four apparatus carry experiments in both tiers, which makes them a controlled comparison of the two; within each only the procedure changes: the torsion table (\#5\,$\to$\,\#13), the circuit panel (\#9\,$\to$\,\#16), the optical rail (\#11\,$\to$\,\#23) and the spectrophotometer (\#30\,$\to$\,\#18\,$\to$\,\#20).

\noindent\textbf{Specification and instances.}
Each experiment is written as one declarative specification of its scene, prior, instruments, gates, milestone checks, tolerance, budget and submission schema.
Its instances are generated from seeds: the hidden parameters are drawn anew for every instance.

\noindent\textbf{Reference solution and tolerances.}
Instruments are modeled with transduction, quantization and noise, and each tolerance is the tightest that they support: a step of 0.2, 0.3, 0.5, 1, 2, 3 or 5\%, or an absolute step where the instrument sets one, limited by resolutions such as the 0.01\,$^\circ$C of a thermometer or the 1\,nm step of a spectrometer scan (\cref{tab:experiments}).
The reference solution meets every tolerance with margin: over its calibration runs, its largest error is 0.23 of the tolerance at the median of the 30 experiments and at most 0.55.

\noindent\textbf{Need for an experiment.}
Across instances the target of a determination task ranges over 6 to about 2900 times its tolerance, 90 times at the median.
A single fixed answer, the best for each experiment, therefore meets the tolerance in 4\% of instances on average, and the designs of the optimization tasks that need no experiment reach the threshold on 6\% (\cref{tab:noexp}).

\section{Protocol}
\label{sec:prereg}
\begin{figure*}[tp]
  \centering
  \includegraphics[width=\linewidth]{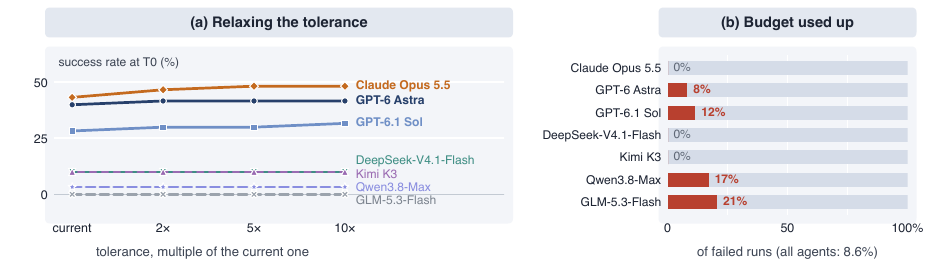}
  \caption{\textbf{Sensitivity to tolerance and budget.} Both panels cover the 30 determination tasks at T0. (a)~Success rate with every scored tolerance multiplied by 1 to 10, all other requirements unchanged. (b)~Failed runs that end with the budget used up, in percent of each agent's failed runs with a recorded end: the simulated time or the wall clock runs out, or the agent stops after using at least 95\% of the simulated time.}
  \label{fig:sensitivity}
\end{figure*}

Every agent writes and runs code against the same public interface and leaves every decision to the model.
GPT-6 Astra, GPT-6.1 Sol and Claude Opus 5.5 run at medium reasoning effort, and DeepSeek-V4.1-Flash, Kimi K3, GLM-5.3-Flash and Qwen3.8-Max with thinking enabled at high reasoning effort, the default of each model; all other settings are default as well.
Across levels only the set of interface tools changes.
A run counts as a failure when it does not succeed: an agent that stops without submitting, an exhausted simulated-time or wall-clock budget, or a submission outside the tolerance or failing a gate.
The process score adds a weight for every milestone a run passes. On a determination task the four milestones that lead to sufficient readings earn 10 each, and knowing the result and restoring the bench 30 each. On an optimization task the last 60 points are divided differently: a valid design earns 20, reaching the threshold of the reference 30, and restoring the bench after a valid design 10.
\Cref{tab:cross} gives the success rate and the process score of every agent at every level on the determination tasks, and \cref{tab:crossopt} on the optimization tasks.
\Cref{tab:cost} gives what a run costs at T0 over all 35 tasks: the agents differ fourfold in tool calls, from 977 to 4009, fivefold in input tokens and more than fifteenfold in output tokens, and 95.4\% to 99.7\% of the input tokens are read from the prompt cache.

\begin{table}[H]
  \centering
  \caption{\textbf{Results by interface level on the determination tasks.} Success rate (\%) with the process score (0--100) in gray at T0, T1 and T2, and the change in success rate between levels in percentage points. T2 is run for four agents (\cref{sec:lab}). The best success rate at each level is in bold.}
  \label{tab:cross}
  \footnotesize\setlength{\tabcolsep}{3pt}
  \newcommand{\ps}[1]{\,{\scriptsize\color{gray}#1}}
  {\begin{tabular}{@{}lccccc@{}}
    \toprule
    Agent & T0 & T1 & T2 & T0$\to$T1 & T1$\to$T2 \\
    \midrule
    Claude Opus 5.5 & \textbf{43.3}\ps{64.0} & \textbf{53.3}\ps{73.7} & -- & $+$10.0 & -- \\
    GPT-6 Astra & 40.0\ps{62.8} & 46.7\ps{61.0} & -- & $+$6.7 & -- \\
    GPT-6.1 Sol & 28.3\ps{55.7} & 30.0\ps{51.7} & -- & $+$1.7 & -- \\
    DeepSeek-V4.1-Flash & 10.0\ps{36.0} & 13.3\ps{36.7} & \textbf{40.0}\ps{57.7} & $+$3.3 & $+$26.7 \\
    Kimi K3 & 10.0\ps{35.7} & 20.0\ps{39.7} & 33.3\ps{58.3} & $+$10.0 & $+$13.3 \\
    Qwen3.8-Max & 3.3\ps{16.3} & 3.3\ps{20.3} & 13.3\ps{35.5} & $+$0.0 & $+$10.0 \\
    GLM-5.3-Flash & 0.0\ps{16.3} & 3.3\ps{20.7} & 13.3\ps{38.3} & $+$3.3 & $+$10.0 \\
    \bottomrule
  \end{tabular}}
\end{table}

\begin{table}[H]
  \centering
  \caption{\textbf{Results by interface level on the optimization tasks.} Success rate (\%) with the process score (0--100) in gray, and the mean attainment, at T0, T1 and T2. T2 is run for the same four agents as in \cref{tab:cross}. The best value in each column is in bold.}
  \label{tab:crossopt}
  \footnotesize\setlength{\tabcolsep}{3pt}
  \newcommand{\ps}[1]{\,{\scriptsize\color{gray}#1}}
  {\begin{tabular}{@{}lcccccc@{}}
    \toprule
    & \multicolumn{2}{c}{T0} & \multicolumn{2}{c}{T1} & \multicolumn{2}{c@{}}{T2} \\
    \cmidrule(lr){2-3}\cmidrule(lr){4-5}\cmidrule(l){6-7}
    Agent & Success & Attain. & Success & Attain. & Success & Attain. \\
    \midrule
    Claude Opus 5.5 & \textbf{30.0}\ps{67.0} & \textbf{0.55} & \textbf{40.0}\ps{76.0} & \textbf{0.63} & -- & -- \\
    GPT-6 Astra & 10.0\ps{61.0} & 0.30 & 20.0\ps{64.0} & 0.28 & -- & -- \\
    GPT-6.1 Sol & 10.0\ps{43.0} & 0.26 & 20.0\ps{59.0} & 0.25 & -- & -- \\
    DeepSeek-V4.1-Flash & 0.0\ps{30.0} & 0.08 & 0.0\ps{34.0} & 0.09 & \textbf{10.0}\ps{40.0} & 0.11 \\
    Kimi K3 & 0.0\ps{16.0} & 0.03 & 0.0\ps{24.0} & 0.05 & \textbf{10.0}\ps{44.0} & \textbf{0.18} \\
    Qwen3.8-Max & 0.0\ps{25.0} & 0.07 & 0.0\ps{30.0} & 0.08 & 0.0\ps{34.0} & 0.10 \\
    GLM-5.3-Flash & 0.0\ps{16.0} & 0.00 & 0.0\ps{20.0} & 0.00 & 0.0\ps{30.0} & 0.00 \\
    \bottomrule
  \end{tabular}}
\end{table}

\begin{table}[t]
  \centering
  \caption{\textbf{Cost of a run at T0.} Tool calls issued by the model, simulated time used, and input and output tokens per run over all 35 tasks, averaged as in \cref{sec:setup}. Cached: share of the input tokens read from the prompt cache. Runs without a usage record are left out.}
  \label{tab:cost}
  \footnotesize\setlength{\tabcolsep}{3pt}
  {\begin{tabular}{@{}lccccc@{}}
    \toprule
    Agent & \makecell{Tool\\calls} & \makecell{Sim.\ time\\(s)} & \makecell{Input\\(M)} & \makecell{Cached\\(\%)} & \makecell{Output\\(k)} \\
    \midrule
    Claude Opus 5.5 & 1824 & 756 & 31.5 & 95.4 & 151.4 \\
    GPT-6 Astra & 977 & 596 & 20.2 & 96.8 & 31.3 \\
    GPT-6.1 Sol & 1146 & 746 & 33.5 & 95.6 & 53.2 \\
    DeepSeek-V4.1-Flash & 1203 & 492 & 52.3 & 99.7 & 257.4 \\
    Kimi K3 & 2516 & 754 & 102.5 & 98.0 & 301.8 \\
    Qwen3.8-Max & 4009 & 671 & 33.5 & 97.0 & 523.2 \\
    GLM-5.3-Flash & 2500 & 539 & 60.6 & 96.5 & 243.7 \\
    \bottomrule
  \end{tabular}}
\end{table}

\section{Sensitivity to tolerance and budget}
\label{sec:sensitivity}
Both analyses cover the 30 determination tasks and rescore their runs of \cref{tab:main} offline, with the success rate averaged as in \cref{sec:setup}.
For the tolerance, every scored tolerance is multiplied by 2, 5 or 10, while the evidence, the consistency of the submission, the bench state, the budget and the single submission are judged as before.
The ranking is the same at every multiple, and no agent's success rate rises by more than 5.0 points (\cref{fig:sensitivity}a).
Of the runs that submit a numerical result, 61.6\% lie within the tolerance; those that fail miss another requirement, such as the evidence or the bench state.
For the budget, a failed run ends with the budget used up if the simulated time or the wall clock runs out or the agent stops after using at least 95\% of the simulated time; this holds for 8.6\% of the failed runs with a recorded end, and for none of those of Claude Opus 5.5, DeepSeek and Kimi (\cref{fig:sensitivity}b).
The other failed runs submit an answer or stop with budget left.

\section{Removing one side of acting to know}
\label{sec:ablationsetup}
The comparison of \cref{tab:ablation} runs on five determination tasks, one from each discipline: Hooke's law (\#1), specific heat (\#24), lens focal length (\#11), black-box resistor (\#9) and strong-acid titration (\#19).
Every condition starts from the run at T0 and removes one side.
The goal, the instances, the tolerance, the milestones, the budget and the scoring stay the same, and the bench must still be restored.

\noindent\textbf{Method given.}
A laboratory handout is appended to the task card.
It states the principle, what to measure and under which conditions, and the calculation with its formulas.
It contains no value of the instance, no position on the bench and no control sequence.
The agent still carries out every operation with the robot and reads every instrument through the cameras.

\noindent\textbf{Operations executed.}
The gripper holds nothing, and the agent works the apparatus through three tools: one moves a part to a place it names, one presses a control, one slides a rider along its rail.
An operation that is physically possible always succeeds: the part is set down at rest where the agent named, or the control is pressed once, and the physics runs on as usual.
The tool returns no reading, no image and no report of the bench.
The agent still decides every operation, observes the feedback through the cameras, for which the arm can still point the wrist camera, and puts every part back.
Each operation is charged simulated time.

\noindent\textbf{Both.}
The handout and the operation tools are given together, and what is left to the agent is reading the instruments.

\noindent\textbf{Readings given.}
This condition covers 27 determination tasks, and there is no bench.
The agent receives the task card and the instrument readings that the reference solution took on the instance: display values, scale readings and exported records, with no method, fit or intermediate result.
It computes the result with code and submits it under the same tolerance.
The success rate is 100\% for the Kimi agent and 96\% for GPT-6 Astra, GPT-6.1 Sol and the DeepSeek, GLM and Qwen agents, against 0\% to 48\% on the same tasks at T0.

\noindent\textbf{Results by task.}
\Cref{tab:ablationtasks} gives the success rate of every task.
With the operations executed, the failures that remain are on two tasks.
On the lens, the DeepSeek and Kimi agents take their readings from an image that is out of focus or misread the position of a rider.
On the specific heat, results lie outside the tolerance of 1\%, with the heat loss left uncorrected.
With the method also given, the agents act and observe as often as without it, with 60 against 56 operations and 34 against 30 observations per run on average, and the further failures are of the same kinds: a pointer read before the tray came to rest, an ammeter misread by a factor of ten, a result outside the tolerance.

\begin{table}[t]
  \centering
  \caption{\textbf{Per-task results of the ablation.} Success rate (\%) on each of the five tasks of \cref{tab:ablation}, under its four conditions that use the bench. The last column is the mean over the five tasks.}
  \label{tab:ablationtasks}
  \footnotesize\setlength{\tabcolsep}{3pt}
  {\begin{tabular}{@{}llcccccc@{}}
    \toprule
    Condition & Agent & \#1 & \#24 & \#11 & \#9 & \#19 & Mean \\
    \midrule
    T0 & GPT-6 Astra & 100 & 0 & 0 & 0 & 0 & 20 \\
     & DeepSeek-V4.1-Flash & 0 & 0 & 0 & 0 & 0 & 0 \\
     & Kimi K3 & 0 & 0 & 0 & 0 & 0 & 0 \\
    \midrule
    Method given & GPT-6 Astra & 100 & 0 & 0 & 0 & 0 & 20 \\
     & DeepSeek-V4.1-Flash & 0 & 0 & 0 & 0 & 0 & 0 \\
     & Kimi K3 & 0 & 0 & 0 & 0 & 0 & 0 \\
    \midrule
    Operations executed & GPT-6 Astra & 100 & 50 & 100 & 100 & 100 & 90 \\
     & DeepSeek-V4.1-Flash & 100 & 100 & 0 & 100 & 100 & 80 \\
     & Kimi K3 & 100 & 50 & 0 & 100 & 100 & 70 \\
    \midrule
    Both & GPT-6 Astra & 100 & 50 & 50 & 100 & 100 & 80 \\
     & DeepSeek-V4.1-Flash & 100 & 50 & 0 & 100 & 100 & 70 \\
     & Kimi K3 & 50 & 0 & 0 & 50 & 100 & 40 \\
    \bottomrule
  \end{tabular}}
\end{table}

\begin{table*}[tp]
  \centering
  \caption{\textbf{Interface levels in full.} What the agent receives at each level. Each level adds one layer to the level below, and tools outside a level are neither listed nor callable.}
  \label{tab:exposure}
  \tablestyle
  \begin{tabularx}{\linewidth}{@{}lL@{}}
    \toprule
    Level & What the agent receives \\
    \midrule
    T0 & Three RGB views with fixed camera calibration, robot state with the hand pose, seven-joint position targets (one per physics step) and gripper commands, an inverse-kinematics solver that turns a hand pose into joint angles without moving the robot, checking collisions or planning a path, the instruments of the apparatus, and \texttt{submit}. Trajectories, collision avoidance, perception and every manipulation skill are the agent's own. \\
    T1 & T0 plus perception and planning: point-prompted segmentation~\cite{sam3} with a zoomed view; ideal depth, rendered without noise and quantized to 1\,mm, with which a chosen pixel is back-projected to the 3D point of the visible surface and a mask to its visible-surface point cloud; Cartesian hand motion and return-to-start planned collision-free~\cite{curobo} against the bench's own collision model (as-built geometry, current part poses and the held object); and a slow, short contact move without collision checking. Object poses, centers and occluded geometry are not given; the agent infers them from the visible surface. \\
    T2 & T1 plus a library of manipulation skills in the agent's workspace, extracted from the successful runs of the GPT and Claude agents: locate, grasp, carry, place, press, push and slide. Each skill is a short sequence of public T1 calls, keeps the hand pointing down and reads only the robot state and the public calibration; the agent may read, copy and change it. \\
    \bottomrule
  \end{tabularx}
\end{table*}

\begin{table*}[tp]
  \centering
  \caption{\textbf{Public tools by group and level.} The tools an agent can call at T0 and those that T1 adds. T2 adds no tools: its skills are code in the agent's workspace that calls the T1 tools.}
  \label{tab:tools}
  \tablestyle\footnotesize
  \begin{tabularx}{\linewidth}{@{}llL@{}}
    \toprule
    Level & Group & Tools \\
    \midrule
    T0 & Task and observation & \texttt{get\_task\_spec} (task card, submission schema, public calibration); \texttt{get\_observation} (the three RGB views, with an evidence id where readings are cited); \texttt{get\_robot\_state} (joint positions, gripper width, whether the gripper holds a part, and the hand pose) \\
       & Motion & \texttt{execute\_joint\_trajectory} (seven-joint position targets, one per physics step); \texttt{set\_gripper}; \texttt{wait} (at most 30\,s of simulated time); \texttt{solve\_ik} (kinematics only, without collision checking) \\
       & Instrument records & one export per experiment that records a dynamic curve, 19 in all: \texttt{export\_logger} (six circuit experiments), \texttt{export\_recording} (\#3, \#6, \#7), \texttt{export\_photogate} (\#5, \#13), \texttt{export\_titration\_log} (\#19, \#21), \texttt{export\_spectro\_scans} (\#20, \#30), \texttt{export\_encoder} (\#4), \texttt{export\_flight} (\#14), \texttt{export\_spectro\_kinetics} (\#22), \texttt{export\_temperature\_log} (\#24); a static reading is read from the cameras \\
       & Submission & \texttt{submit\_result} (once; the first submission closes the episode) \\
    \midrule
    T1 & Perception & \texttt{segment\_sam3\_point\_prompt} (mask candidates at a chosen pixel); \texttt{inspect\_mask\_image} (close-up of a candidate); \texttt{localize\_rgbd\_point} (3D point of a chosen pixel); \texttt{mask\_geometry} (bounds and median of a mask's visible points) \\
       & Planning & \texttt{move\_hand} (collision-free path to a hand pose); \texttt{plan\_return\_to\_start}; \texttt{translate\_contact} (slow straight move of at most 5\,cm, to touch or back out of a contact); \texttt{get\_robot\_pose} \\
    \bottomrule
  \end{tabularx}

\end{table*}

\section{Interface reference}
\label{sec:api}
\Cref{tab:exposure} gives what the agent receives at each interface level, and \cref{tab:tools} the public tools of each level.

\section{Experiment list}
\label{sec:cards}
Each experiment consists of a task card for the agent and a declarative specification for the evaluator. \Cref{tab:card} condenses both for the cantilever (\#15), the experiment of \cref{fig:build}; the cards and specifications of all 35 tasks will be released with the benchmark.

\begin{table*}[tp]
  \centering
  \caption{\textbf{The 30 determination tasks.} Target, required evidence, tolerance, discipline, capability dimension and tier of each task. Tol.: tolerance, relative unless a unit is given; Dim.: primary capability dimension. Letters mark the decisions an advanced experiment requires (\cref{sec:details}): (a)~choosing the conditions, (b)~sequential design, (c)~model identification, (d)~acting on the result. Evidence: what a run must contain beyond readings cited from its own episode that match the instrument; $^\ast$: the result must also agree with the evaluator's refit of the cited readings.}
  \label{tab:experiments}
  \tablestyle\footnotesize\setlength{\tabcolsep}{3.5pt}\renewcommand{\arraystretch}{0.95}
  \begin{tabularx}{\linewidth}{@{}rl>{\hsize=.8\hsize}L>{\hsize=1.2\hsize}Lllll@{}}
    \toprule
    \# & Experiment & Target & Evidence the run must contain & Tol. & Disc. & Dim. & Tier \\
    \midrule
    1 & Hooke's law & spring stiffness $k$ & ruler readings at $\ge$4 loads, tray at rest and untouched$^\ast$ & 2\% & Mech. & Read & basic \\
    2 & Pendulum $g$ & acceleration $g$ & a completed timing of each of P1--P3, read from its timer$^\ast$ & 0.3\% & Mech. & Manip. & basic \\
    3 & Coupled oscillators & normal-mode frequencies $\omega_1<\omega_2$ & recordings with an untouched stretch of $\ge$40\,s showing both modes & 0.2\% & Mech. & Manip. & basic \\
    4 & Spring oscillation & spring stiffness $k$ from oscillation & encoder records of $\ge$5 free periods at $\ge$3 loads$^\ast$ & 0.2\% & Mech. & Data & basic \\
    5 & Torsion pendulum & moment of inertia of an unknown body & photogate records of the empty turntable, the disk and the body & 1\% & Mech. & Data & basic \\
    6 & Air-track collision & mass of the unknown cart & a recorded collision, B at rest and no outside force on either cart$^\ast$ & 0.5\% & Mech. & Data & basic \\
    7 & Incline friction & friction coefficients $\mu_s$, $\mu_k$ & records of each slider starting to slip and sliding freely & 5\%, 0.5\% & Mech. & Manip. & basic \\
    8 & RC capacitance & capacitance $C$ & meter or logger readings with the box in a closed circuit or charged & 0.5\% & Elec. & Data & basic \\
    9 & Black-box resistor & resistance $R$ & meter or logger readings with the supply on and the box in circuit & 0.3\% & Elec. & Data & basic \\
    10 & LED wavelength & peak wavelength of an unknown LED & settled readings of each LED used, the unknown one conducting & 2\% & Elec. & Long & basic \\
    11 & Lens focal length & focal length $f$ & rail readings with the lens seated and the image in focus & 1\% & Opt. & Read & basic \\
    12 & Grating wavelength & laser wavelength $\lambda$ & rail and spot positions with both first orders on the screen & 1\% & Opt. & Read & basic \\
    25 & Single-slit width & slit width $a$ & dark-fringe positions on both sides of the central maximum & 1\% & Opt. & Read & basic \\
    26 & Plate index & refractive index $n$ & focused images with and without the plate & 3\% & Opt. & Read & basic \\
    27 & Battery EMF & EMF $E$ and internal resistance $r$ & settled meter or logger readings, nothing switched during them & 0.5\% & Elec. & Data & basic \\
    28 & Thermistor $B$ & B constant and $R_{25}$ & display readings with the probe in the block's well & 0.3\% & Therm. & Data & basic \\
    29 & Thermal expansion & expansion coefficient $\alpha$ & display readings of temperature and extension & 0.5\% & Therm. & Read & basic \\
    30 & Dye absorptivity & absorption peak $\lambda_{\max}$ and absorptivity $\varepsilon$ & absorbance readings of the stocks or their dilutions & 1\,nm, 2\% & Chem. & Manip. & basic \\
    \midrule
    13 & Centroid \& inertia & center of mass $x_{\mathrm{com}}$ and $I_{\mathrm{com}}$ & photogate records of the body on $\ge$2 holes, with references & 0.3\,mm, 0.5\% & Mech. & Long & adv.\ (c) \\
    14 & Spring launcher & landing distance at a set notch (launched) & complete shot records free of the robot; no test shot at the set notch & 2\% & Mech. & Manip. & adv.\ (d) \\
    15 & Cantilever & bending stiffness $EI$ and clamp stiffness $k_\theta$ & indicator readings with the carriage at rest in a hole & 0.5\% & Mech. & Read & adv.\ (a,c) \\
    16 & 3-terminal black box & network topology and $R_1$, $R_2$, $C_1$ & meter or logger readings with the box in a closed circuit or charged & 0.5\% & Elec. & Long & adv.\ (a,c) \\
    17 & Rectifier supply & $R_{\min}$ of three filter modules; final supply & logger records of the modules and of the final supply under load & 1\% & Elec. & Long & adv.\ (a,d) \\
    18 & Spectro calibration & concentration of an unknown dye & absorbance readings, at least one of the unknown & 2\% & Chem. & Long & adv.\ (a,b) \\
    19 & Strong-acid titration & HCl concentration & recorded pH rows with the electrode in a sample on the stirrer & 0.5\% & Chem. & Manip. & adv.\ (b) \\
    20 & Two-dye mixture & both concentrations in a two-dye mixture & absorbance readings, at least one of the mixture & 3\% & Chem. & Long & adv.\ (a,c) \\
    21 & Weak-acid titration & weak-acid concentration and p$K_a$ & recorded pH rows with the electrode in a sample on the stirrer & 1\%, 0.03 & Chem. & Manip. & adv.\ (b,c) \\
    22 & Crystal-violet kinetics & second-order rate constant $k$ & kinetic runs of two cuvettes at different [OH$^-$] & 2\% & Chem. & Data & adv.\ (a,c) \\
    23 & Two-lens relay & upright image at a set magnification (verified) & one observation with both lenses seated and the image in focus & 3\% & Opt. & Long & adv.\ (c,d) \\
    24 & Specific heat & specific heat $c_s$ & recorded heater and water temperatures before and after mixing & 1\% & Therm. & Data & adv.\ (b,c) \\
    \bottomrule
  \end{tabularx}
\end{table*}

\section{Failure causes}
\label{sec:causes}
\begin{figure*}[tp]
  \centering
  \includegraphics[width=0.93\linewidth]{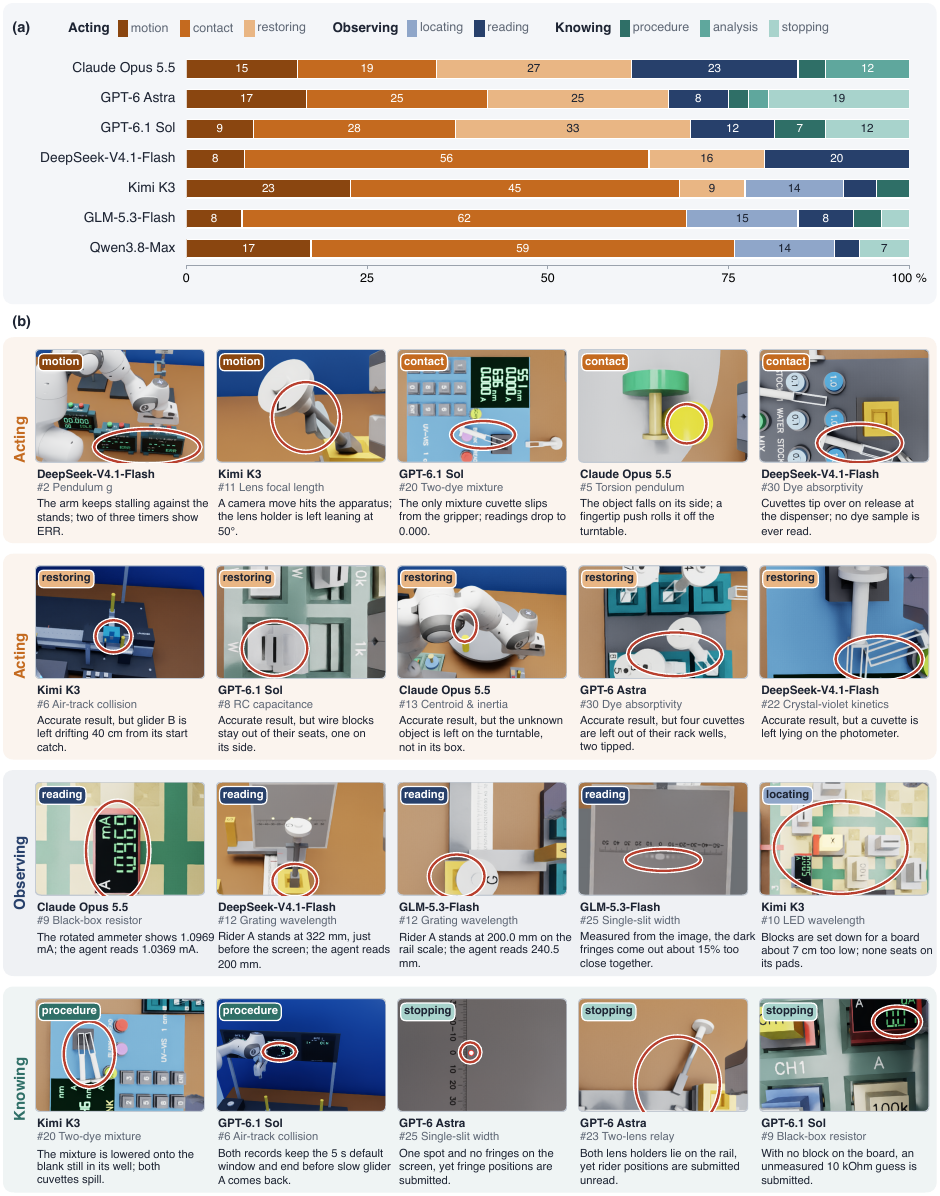}
  \caption{\textbf{Failure causes at T0.} (a) Failed runs by the cause of their first decisive error, in percent of each agent's failed runs with an attributed cause, grouped by the stage of acting to know. (b) Typical failures of each stage, each shown in a frame from the agent's own cameras with the decisive spot ringed; the chip on each frame names the cause.}
  \label{fig:failures}
\end{figure*}

A failed run is attributed to its first decisive error: the earliest error, not corrected by the agent, after which the run can no longer succeed.
Errors the agent corrects do not count, and the last visible symptom is not used.
When two causes are present, the earlier one that alone suffices for failure is taken.
A run that reaches L5 and fails only at restoring the bench is attributed to acting: in every such run a part was put back out of place or knocked over.
\Cref{fig:failures}a gives the shares of the causes among each agent's failed runs with an attributed cause, and \cref{fig:failures}b shows typical failures of each stage, from mechanics, optics, circuit and chemistry benches: the arm colliding with the apparatus, parts slipping or tipping in the gripper, parts put back out of place after an accurate result, digits, rail scales and fringe positions read wrongly, a board height misjudged, a procedure cut short, and results submitted without a measurement.

\begin{table*}[tp]
  \centering
  \caption{\textbf{Coverage by capability dimension and tier.} The 30 determination tasks by primary capability dimension and tier, with the number of tasks in parentheses. Superscript numbers mark same-apparatus pairs (\cref{sec:details}); targets and tolerances are in \cref{tab:experiments}, benches in \cref{fig:gallery}.}
  \label{tab:design}
  \tablestyle\footnotesize
  \begin{tabularx}{\linewidth}{@{}l>{\hsize=1.08\hsize}L>{\hsize=.92\hsize}L@{}}
    \toprule
    Capability dimension & Basic tier (18) & Advanced tier (12) \\
    \midrule
    Fine manipulation and contact (7) & Pendulum $g$ (\#2), coupled oscillators (\#3), incline friction (\#7), dye absorptivity$^{4}$ (\#30) & Spring launcher (\#14), strong-acid titration (\#19), weak-acid titration (\#21) \\
    Long-horizon multi-stage work (7) & LED wavelength (\#10) & Centroid and inertia$^{1}$ (\#13), three-terminal black box$^{2}$ (\#16), rectifier supply (\#17), spectrophotometric calibration$^{4}$ (\#18), two-dye mixture$^{4}$ (\#20), two-lens relay$^{3}$ (\#23) \\
    Fine reading (7) & Hooke's law (\#1), lens focal length$^{3}$ (\#11), grating wavelength (\#12), single-slit width (\#25), plate refractive index (\#26), thermal expansion (\#29) & Cantilever compliance (\#15) \\
    Multi-point acquisition and analysis (9) & Spring oscillation (\#4), torsion pendulum$^{1}$ (\#5), air-track collision (\#6), RC capacitance (\#8), black-box resistor$^{2}$ (\#9), battery EMF (\#27), thermistor $B$ (\#28) & Crystal-violet kinetics (\#22), specific heat (\#24) \\
    \bottomrule
  \end{tabularx}
\end{table*}

\begin{table*}[tp]
  \centering
  \caption{\textbf{One experiment in full.} The cantilever experiment (\#15), condensed from its task card for the agent and its specification for the evaluator.}
  \label{tab:card}
  \tablestyle\footnotesize
  \begin{tabularx}{\linewidth}{@{}>{\bfseries}p{2.6cm}L@{}}
    \toprule
    Question & Determine the bending stiffness $EI$ of a metal strip clamped at one end and the rotational stiffness $k_\theta$ of its clamp, which turns in proportion to the moment at the strip's root. \\
    Apparatus & The strip sticks out horizontally from its clamp; four knife-edge saddles ride on it at printed distances from the clamp; three steel weights of 50, 100 and 100\,g stand in a rack. A digital indicator on a carriage reads the strip's height on an LCD in mm to three decimals; the carriage clicks into holes along a rail, each printed with its distance from the clamp, and a \texttt{ZERO} cap sets the reading to zero where it stands. \\
    Hidden quantities & $EI\sim\mathcal{U}(3,8)$\,N\,m$^2$ and $k_\theta\sim\mathcal{U}(80,300)$\,N\,m/rad, scored; the indicator offset $\mathcal{U}(-0.5,0.5)$\,mm, the rail tilt $\mathcal{U}(-1.5,1.5)$\,mm/m and the bow of the unloaded strip $\mathcal{U}(-0.3,0.3)$\,mm, which make the unloaded reading differ from hole to hole. \\
    Tier & Advanced: (a)~the agent chooses which loads at which distances to read, and (c)~$EI$ and $k_\theta$ add to every deflection and are identifiable only jointly, from readings whose load moments differ. \\
    Dimension & Fine reading. \\
    Acceptance & Every cited reading comes from a different observation of the episode, with the carriage at rest in the named hole and the value equal to what the LCD showed within 0.001\,mm. At submission all weights stand in the rack, nothing rests on the strip or its saddles, the carriage stands in a hole, the robot holds nothing, and no loose item is tipped over or off the bench. One submission within the budget. \\
    Process checks & L1: a weight lifted, \texttt{ZERO} pressed or the carriage pushed. L2: a weight pressing on the strip, untouched by the robot, with the carriage in a hole, held for 1.2\,s. L3: a first observation of such a loaded state with the LCD readable. L4: valid readings whose differences at a common hole span both unknowns. L5 and L6 as for every experiment. \\
    Tolerance & 0.5\% on each of $EI$ and $k_\theta$ (the larger relative error is scored). \\
    Reference error & At most 0.23\% (0.46 of the tolerance) over its calibration runs. \\
    \bottomrule
  \end{tabularx}
\end{table*}

\end{document}